\documentclass[letterpaper, 10 pt, twocolumn, conference, numbers,sort&compress]{ieeeconf}

\IEEEoverridecommandlockouts                              

\let\labelindent\relax
\usepackage{graphicx}
\usepackage{booktabs}
\usepackage{cite}
\usepackage{amsmath}
\usepackage{algorithm}
\usepackage{algorithmic}
\usepackage{amssymb}
\usepackage{lipsum}
\usepackage{caption}
\usepackage{xcolor}
\usepackage{placeins}
\usepackage{booktabs} 
\usepackage{pifont}
\usepackage{multirow}
\usepackage{subfig}
\usepackage{hyperref}
\hypersetup{
    colorlinks = true,
    linkcolor = blue,
    urlcolor = blue,
}
\usepackage{enumitem}
\usepackage[T1]{fontenc}
\usepackage{CJKutf8}
\usepackage{url}

\newcommand{\nop}[1]{} 
\newcommand{\mode}{eVTOL aircraft}

\title{\LARGE \bf
TEeVTOL: Balancing Energy and Time Efficiency in eVTOL Aircraft Path Planning Across City-Scale Wind Fields
}

\author{Songyang Liu$^{1}$, Shuai Li$^{1}$, Haochen Li$^{1}$, Weizi Li$^{2}$, Jindong Tan$^{3}$
\thanks{$^{1}$Songyang Liu, Shuai Li, Haochen Li are with Department of Civil and Environmental Engineering at University of Tennessee, Knoxville, TN, USA {\tt\small sliu78@vols.utk.edu; \{sli48, hli111\}@utk.edu}}%
\thanks{$^{2}$Weizi Li is with Min H. Kao Department of Electrical Engineering and Computer Science at University of Tennessee, Knoxville, TN, USA {\tt\small weizili@utk.edu}}%
\thanks{$^{3}$Jindong Tan is with Department of Mechanical, Aerospace and Biomedical Engineering at University of Tennessee, Knoxville, TN, USA {\tt\small tan@utk.edu}}%
}
\begin{document}
\begin{CJK}{UTF8}{gbsn}

\maketitle
\thispagestyle{empty}
\pagestyle{empty}

\begin{abstract}
Electric vertical-takeoff and landing (eVTOL) aircraft, recognized for their maneuverability and flexibility, offer a promising alternative  to our transportation system.
However, the operational effectiveness of these aircraft faces many challenges, such as the delicate balance between energy and time efficiency, stemming from unpredictable environmental factors, including wind fields. 
Mathematical modeling-based approaches have been adopted to plan aircraft flight path in urban wind fields with the goal to save energy and time costs.
While effective, they are limited in adapting to dynamic and complex environments.
To optimize energy and time efficiency in \mode{}'s flight through dynamic wind fields, we introduce a novel path planning method leveraging deep reinforcement learning. 
We assess our method with extensive experiments, comparing it to Dijkstra's algorithm—the theoretically optimal approach for determining shortest paths in a weighted graph, where weights represent either energy or time cost.
The results show that our method achieves a graceful balance between energy and time efficiency, closely resembling the theoretically optimal values for both objectives.

\end{abstract}


\section{Introduction}
Urban air mobility (UAM) is an efficient transportation system where everything from small package-delivery drones to passenger-carrying air taxis operate over urban areas~\cite{Kelsey}. 
It provides an alternative to current ground transportation systems by unlocking traffic supply in low-altitude urban spaces~\cite{straubinger2020overview,cohen2021urban,bauranov2021designing}.  
The demand for UAM is currently under active exploration and development by various stakeholders, for example, NASA and Uber are establishing aerial transport systems for both packages and passengers~\cite{Abby}.
The UAM market is projected to reach \$1.5 trillion by 2040~\cite{Alan}, which  includes autonomous flying cars, electric vertical-takeoff and landing (eVTOL) aircraft, and personal air vehicles.
The potential growth has garnered an even more optimistic projection of \$2.9 trillion, underscoring the expectations and confidence in the evolution and adoption of UAM~\cite{Alan,Paul2022}. 

One of the most promising modes of UAM is the \mode{}, which is 
expected to enhance urban logistics, emergency services, and more. 
A central task of \mode{} is path planning that optimizes flight paths for energy and time efficiency 
in an urban environment filled with static and dynamic objects such as buildings and other UAM vehicles~\cite{Ware2016,Wu2021,Zhou2020}. 
In contrast to ground vehicles, which are constrained by roads and other vehicles, \mode{} enjoy a multitude of flight path options. 
Nevertheless, diverse flight paths would result in varying levels of energy consumption and flight times. 
Furthermore, intricate wind fields produced by urban structures and terrain~\cite{Ware2016}, and other unpredictable environment factors complicate the planning and optimization of \mode{}'s flight paths~\cite{Hong2021}. 

\begin{figure}
    \centering
    \includegraphics[width=\linewidth]{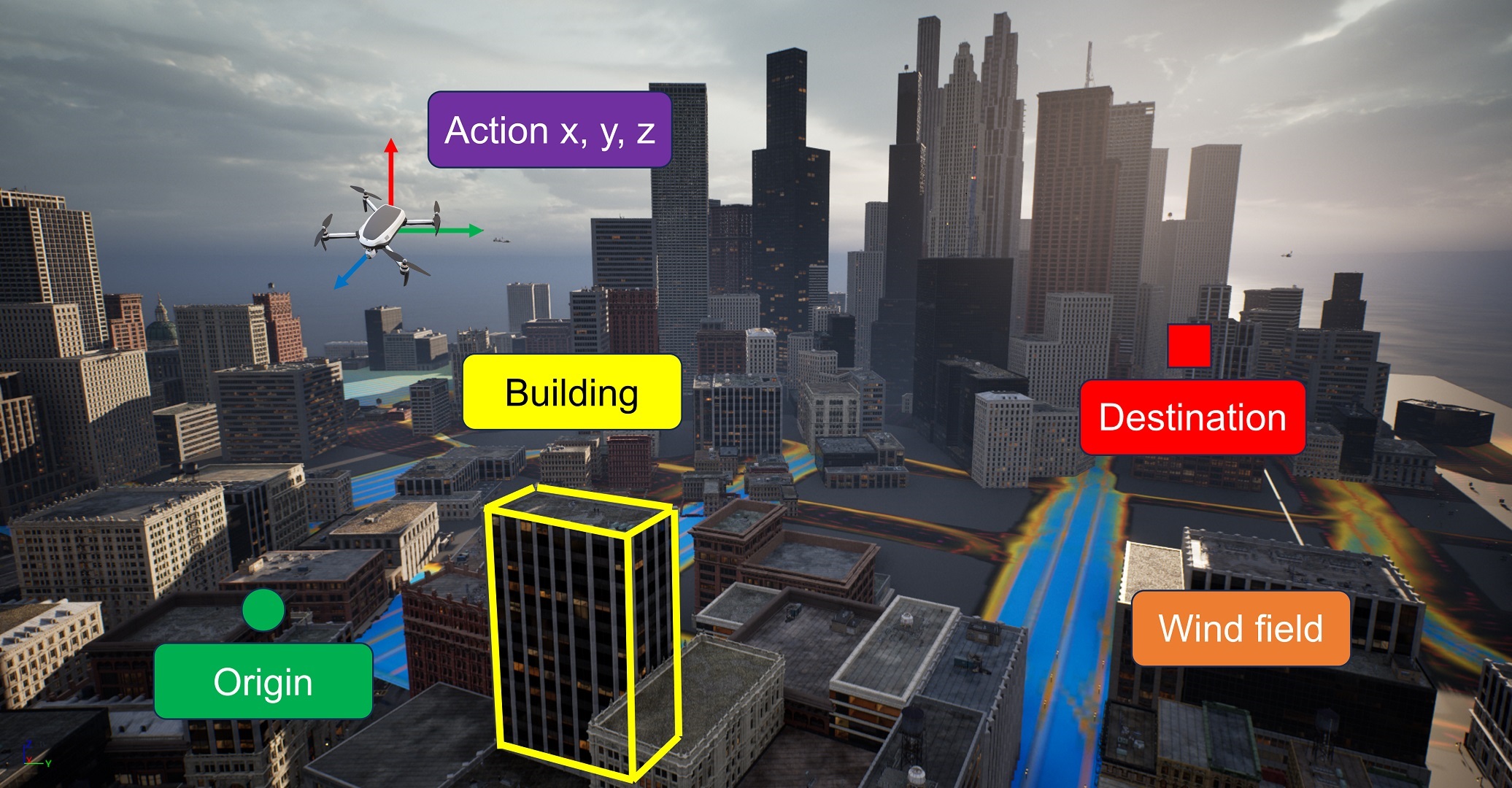}
    \caption{Path planning of eVTOL aircraft through city-scale wind fields enabled by deep reinforcement learning. For illustration purposes, the wind field is shown in 2D. }
    \label{fig:teaser}
    \vspace{-1.5em}
\end{figure}


Mathematical modeling has been employed for planning aircraft flight paths~\cite{Forkan, Chen}. However, these methods often grapple with the intricacies of risk assessment and multi-objective optimization, particularly in densely populated areas~\cite{Babu2022,Wang2023}, and struggle to swiftly adapt to environment changes. 
In contrast, machine learning-based methods offer adaptability, robustness, and efficiency. They excel in handling dynamic and uncertain environments, adapting to real-time changes, and are more scalable for complex urban scenarios~\cite{ramezani2023uav, tu2023uav, maciel2019online}. 
In \mode{} flight, there is a key trade-off between energy consumption and flight time:
reducing flight time, particularly in headwind conditions, often increases energy use due to the aircraft's acceleration or direct routing, while energy conservation strategies, such as leveraging downwind routes, may extend flight duration. 
Thus, it is challenging to balance these goals and ensure efficient destination reach without excessive energy drain.

We introduce a novel path planning technique for \mode{} navigating through city-scale wind fields. 
Our approach, enabled by deep reinforcement learning, adeptly balances the energy conservation and time efficiency of \mode{} in highly dynamic environments.
Specifically, we integrate the energy and time costs associated with \mode{} flight into the training process. 
To learn the optimal policy for the aircraft's path planning, we adopt Proximal Policy Optimization (PPO)~\cite{Schulman2017} and significantly tailor it to our purpose.  
For reward design, we distinguish between non-terminating reward and terminating reward, taking into account diverse scenarios of \mode{} flight within urban environments to enhance the efficacy of learning.  
Additionally, we adopt curriculum learning~\cite{bengio2009curriculum} by dividing the training of \mode{} into several stages based on its flying distance to facilitate learning.
To evaluate the performance, we compare our method with Dijkstra's algorithm, the theoretical optimal method on finding the shortest paths in a weighted (energy or time) graph.
We conduct comprehensive experiments under various wind fields and original-destination pairs. 
The results show that our method closely aligns with the theoretically optimal values for both energy consumption and time cost, with no statistically significant differences observed. Various path examples are also shown to demonstrate the effectiveness of our method.

\section{Related Work}
There are two major approaches in planning paths for aerial vehicles: mathematical modeling-based and machine learning-based methods. 
The former often employs heuristic algorithms. For example, ant colony optimization is adopted for improving search capabilities in aircraft path planning~\cite{Ahmed,Wan2023,Li2023}. 
Cuckoo search algorithm is applied to design an energy-efficient path in aircraft-enabled wireless networks~\cite{Zhu2021}. 
Chodnicki et al. develop an aircraft mathematical model considering forces and moments~\cite{Chodnicki2022}. 
A mixed-integer linear programming is proposed for path planning of multiple aerial vehicles in urban areas with obstacles of different heights~\cite{Bahabry2019}. 
While mathematical modeling offers precise solutions for aircraft path planning, its high computational demands and complexity often make it impractical for real-time and dynamic applications~\cite{Yao2014,Yang2020,Zhang2020,Bai2021,Liu2021,Sandino2022,Zhou2023}. In contrast, machine learning-based methods offer adaptability and efficiency. They excel in handling dynamic and uncertain environments, and are more scalable for complex urban scenarios~\cite{ramezani2023uav, tu2023uav, maciel2019online}.

Among various machine learning techniques, reinforcement learning (RL) has emerged as an effective tool for aircraft path planning. 
To provide some examples, Xu et al. develop a DQN-based aircraft path planning algorithm to avoid the obstacles~\cite{Xu2022}. 
Li et al. propose a stepwise DQN algorithm to extract common features among different navigation targets~\cite{Li2022}. 
Wang et al. introduce a D3QN-based approach for efficient real-time navigation of aerial vehicles~\cite{Wang2022}. 
Luna et al. demonstrate the DQN's effectiveness in achieving optimal mission coverage~\cite{Luna2022}. 
More recent studies employ Policy Gradient (PG) methods, which offer faster convergence, better adaptability, and improved efficiency in complex and dynamic scenarios.
For example, Deep Deterministic PG (DDPG) is adopted for adjusting the flight altitude of a moving mass–actuated aircraft~\cite{Qiu2022}. 
Twin Delayed DDPG (TD3) and its variants optimize aircraft responses for obstacle avoidance~\cite{Liu2022,Zhang2023,Zhang2023_xuan,Hu2023}. 
To overcome TD3's sensitivity to hyperparameters, PPO has gained popularity for its capacity to mitigate drastic policy changes. The inherent stability and efficiency in policy adjustment make PPO a suitable choice for our task~\cite{Xu2024}.


\section{Methodology}

We begin by explaining our problem formulation and learning techniques, followed by the introduction of wind field simulation. 


\subsection{Energy-aware \mode{} Path Planning} 
We formulate our task as a Partially Observable Markov Decision Process (POMDP) represented by a tuple ($\mathcal{S}$, $\mathcal{A}$, $\mathcal{P}$, $\mathcal{R}$, $\gamma$, $T$, $\Omega$, $\mathcal{O}$) where: $\mathcal{S}$ is the state space; $\mathcal{A}$ is the action space; $\mathcal{P}(s'|s,a)$ is the transition probability function; $\mathcal{R}$ is the reward function; $\gamma\in(0, 1]$ is the discount factor; $T$ is the episode length (horizon); $\Omega$ is the observation space; and $\mathcal{O}$ is the probability distribution of retrieving an observation $\omega \in \Omega$ from a state $s \in \mathcal{S}$. 
At each timestep $t \in [1,T]$, an \mode{} uses its policy $\pi_{\theta}(a_t|o_t)$ to take an action $a_t$ $\in$ $\mathcal{A}$, given the observation $o_t$ $\in$ $\mathcal{O}$. Next, the environment provides feedback on action $a_t$ by calculating a reward $r_t$ and transitioning the agent into the next state $s_{t+1}$. The \mode{}'s goal is to learn a policy $\pi_{\theta}$ that maximizes the discounted sum of rewards, i.e., return, $R_t = \sum^{T}_{i=t}\gamma^{i-t}r_i$. 



\subsubsection{Action space}
The action space consists of three discrete actions \{-1, 0, 1\} along the X-, Y-, Z-axis, respectively.  
When $a_x = 1$, the \mode{} advances to the adjacent cell in the positive X-direction; if $a_x = 0$, the aircraft remains stationary along the X-axis; if $a_x = -1$, the aircraft moves to the adjacent cell in the negative X-direction. The same goes for Y- and Z-axis.
Upon executing an action, the aircraft transitions from the center of one cell to the center of the next cell. 
$$
A  = \{(a_x, a_y, a_z)\}, a_x, a_y, a_z \in \{-1; 0; 1\}. 
$$
  

\subsubsection{Observation space}
To enable an RL policy to generalize across a variety of scenarios, we transform the conditions each \mode{} observes into a fixed-length representation, which includes the following. 

\begin{itemize}[leftmargin=*]
    
\item Position info: We use $c^t$ to represent the cell that the aircraft resides at time $t$. We convert the 3D coordinates of aircraft $p^t$ and the corresponding cell $c^t$ as follows: 
$$
c^t = round\left(\left(p^t-X_{min}\right)|X_{cell}\right),
$$
$$
p^t = \left(c^t+0.5\right)*X_{cell}+X_{min},
$$

where $X_{min}$ is the minimal coordinate along the X-axis. $X_{cell}$ is the side length of each cell along the X-axis. 
Same conversion is pursued for Y- and Z-axis. 
We further define $des$ as the position of the flight destination and $det$ as the distance to the nearest building and environment boundary in six directions: front, back, left, right, up, down: 
$$
det = \{d_{\text{front}}, d_{\text{back}}, d_{\text{left}}, d_{\text{right}}, d_{\text{up}}, d_{\text{down}}\}.
$$

\item Energy consumption: We calculate the energy cost of \mode{} flying with air resistance as follows~\cite{Li2022}: 

$$
E_t = \frac{L}{V\eta_{P} \eta_{M}\eta_{ESC}} \left[ \frac{1}{2} \rho V^3 S C_{D0} + \frac{2k M^2 g^2}{\rho S V} \right], 
$$
where $L$ is flight distance, 
$V$ is the \mode{}'s relative velocity to the air, 
$S$ represents the windward area of the \mode{}, defined as the surface area facing into the direction from which the wind is coming. $C_{D0}$ is the zero-lift drag coefficient, $\rho$ is the air density, $k$ is the induced drag factor, $M$ is the total takeoff mass of the \mode{},
$g$ is the gravity acceleration (i.e., 9.8 $m/s^2$).  
$\eta_{P}$, $\eta_{M}$, and $\eta_{ESC}$ are the efficiency of the propeller, brushless motor, and brushless electronic speed control, respectively.
We denote $E_{rem}$ as the remaining energy percentage of the aircraft.  
 
\item Wind field: The wind vector is defined as: 
$$
    W = (x,y,z,u,v,w),
$$

where $(x,y,z)$ represents a point in the simulation area and $(u,v,w)$ represents the wind velocity at $(x,y,z)$.
\end{itemize} 

\noindent Overall, the observation space of an \mode{} at $t$ is: 
$$
    o^t = \langle c^t \rangle \oplus \langle des \rangle \oplus \langle det \rangle \oplus \langle W \rangle\oplus \langle E_{rem}, E_t, E_{t-1}, E_{t-2} \rangle. 
$$

\subsubsection{Reward function}
Our reward function consists of non-terminating reward $r_{NT}$ (for intermediate steps) and terminating reward $r_{T}$ (for the terminate step). The non-terminating reward function takes the following format:

$$
  r_{NT}  = \alpha_1 E_t + \alpha_2 T + \alpha_3 D_{diff},
$$

\noindent where $D_{diff}$ is the difference between the current distance and the next distance to the destination.
$T$ is the \mode{}'s travel time between cells. 
The weights are set empirically as $\alpha_1$ = -3.5, $\alpha_2$ = -1.25, and $\alpha_3$ = -0.04.  
We terminate training when the \mode{} is \ding{182} out-of-bounds (exiting simulation), \ding{183} in collision, \ding{184} depleting energy, \ding{185} exceeding a predefined time limit, or \ding{186} successfully reach the destination. 
For case \ding{186}, we set $r_T = 1000$, and for all other cases, $r_T = -100$.
   
To ensure the RL agent not only reaches the destination successfully but also continually optimizes the flight path to minimize energy or time costs, we pursue reward shaping: an additional reward adjustment is introduced at the end of each successful episode. If the agent’s energy or time cost in a successful episode is lower than the historical best, it earns additional rewards based on the cost difference; conversely, incurring a cost higher than previous best results in a penalty, which is deducted from the basic success reward. 
This setup encourages agents to continually seek more efficient flight paths. 
This reward mechanism does not apply to the first episode that successfully reaches the destination as there is no historical optimal value for comparison at that time.
This dynamic reward adjustment effectively puts the focus of RL on continuous performance improvement rather than just completing the task itself.





\subsubsection{RL algorithm}
We use PPO~\cite{Schulman2017} to learn the optimal policy. The original PPO paper provided limited implementation details beyond the use of Generalized Advantage Estimation (GAE) for the advantage function calculation. 
The details of neural network architecture or activation function are left unspecified, allowing for customization based on the problem at hand. 
However, as Engstrom et al.~\cite{Engstrom2020} suggest, even superficial or seemingly trivial changes in optimization methods or algorithmic tweaks can significantly impact PPO's performance. Our modification thus considers various factors: we use the tanh activation function; include LayerNorm, BatchNorm, and Dropout layers in both actor and critic networks; and adopt linearly decay learning rate.
We further normalize the reward to mitigate impacts on the value function training caused by excessively large or small rewards. We record the standard deviation of a rolling discounted sum of rewards, $\sigma=std(\sum^{T}_{i=t}\gamma^{i-t}r_i)$, and normalize the current reward as $r_t/\sigma$.

\begin{figure}
    \centering
    \includegraphics[width=\linewidth]{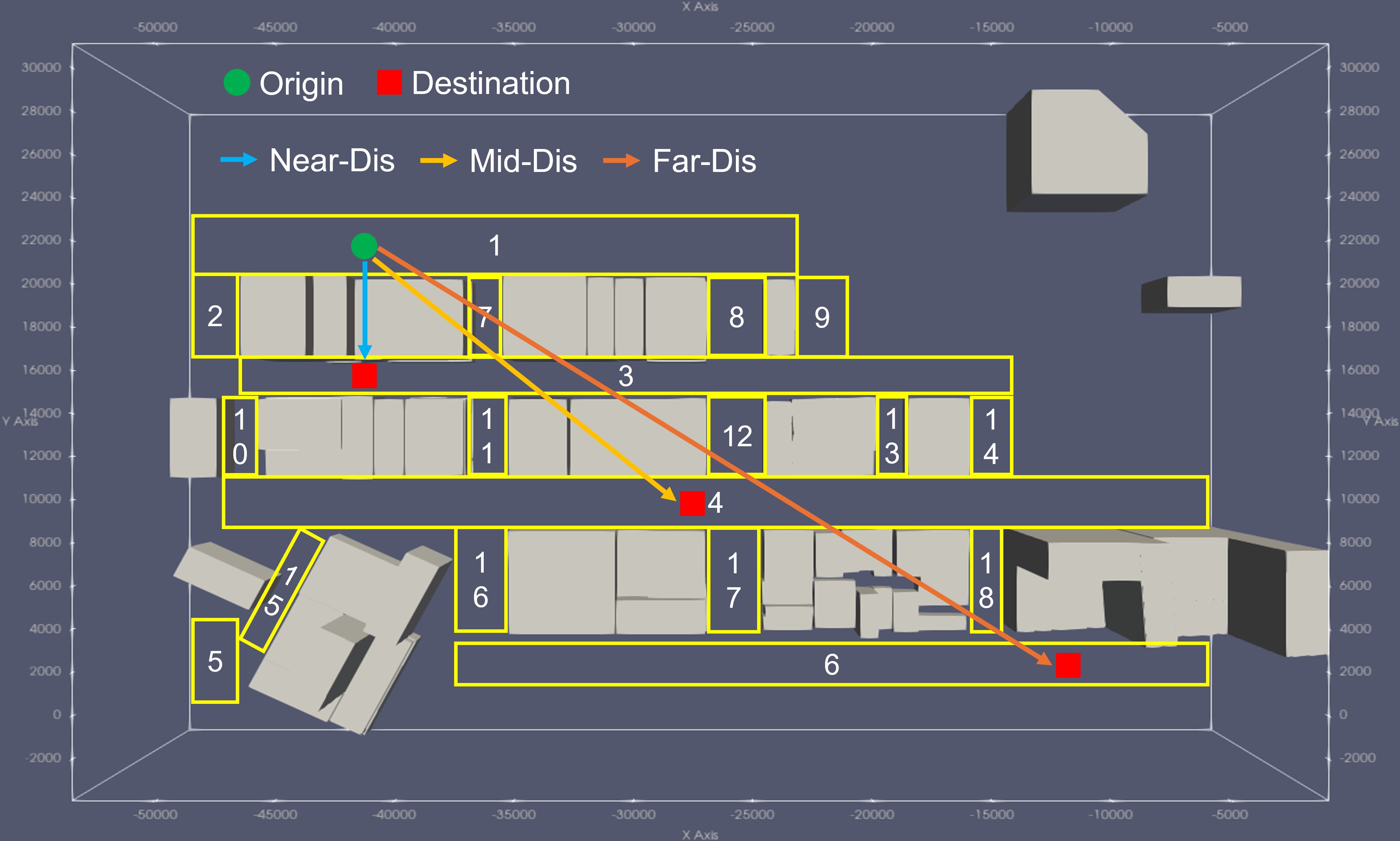}
    \caption{For stage training, we divide the urban environment into 18 areas. Origin-destination pairs are sampled within these areas and are categorized into three classes: near-distance, mid-distance, and far-distance. During training, the \mode{} will learn to master near distances first, then move on to further distances.}
    \label{fig:training}
    \vspace{-1.5em}
\end{figure}

\subsubsection{Stage Training}

We divide the training environment into 18 areas as shown in Fig.~\ref{fig:training}. Various origin-destination pairs are sampled within these areas.  
We categorize the distances between origins and destinations into three classes: near-distance, mid-distance, and far-distance.
To facilitate learning, we employ a curriculum learning approach~\cite{bengio2009curriculum,Shen2022IRL}, sequentially training \mode{} on near-distance, followed by mid-distance, and finally far-distance.





\subsection{Simulation of City-Scale Wind Fields}
\begin{figure*}[ht]
    \centering
    \includegraphics[width=\linewidth]{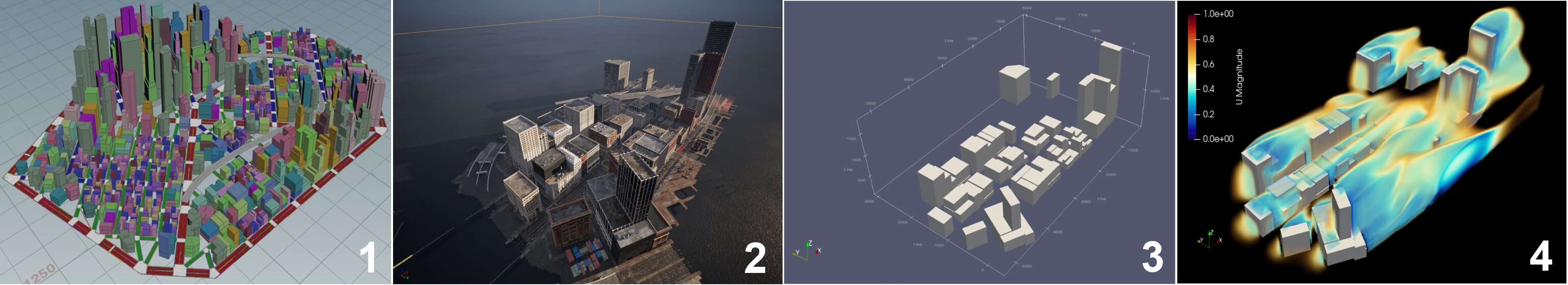}
    \caption{City-scale wind field simulation. 1. We first create a detailed 3D model of an urban area in SideFX: Houdini~\cite{Houdini}. 
    2. The 3D model is then imported into Unreal Engine~\cite{Unreal} to create an immersive simulation environment. 
    3. Next, the simulation environment is imported into OpenFOAM~\cite{OpenFOAM} via scripting. This allows for the visualization of the environment in Paraview~\cite{Paraview}. 
    4. The result of wind field simulation is visualized as volumetric rendering of velocity field magnitude.}
    \label{fig:wind}
    \vspace{-1.5em}
\end{figure*}

To simulate high-fidelity, city-scale wind fields (see Fig.~\ref{fig:wind}), we use Reynolds-averaged incompressible Navier-Stokes equations (RANS)~\cite{alfonsi2009reynolds} to simulate steady-state wind fields. 
The RANS simulations are carried out with an open-source finite-volume method (FVM) code of OpenFOAM \cite{Weller1998}. The RANS equations are defined in Eqs. \ref{eq:continuity} and \ref{eq:RANS}. 

\begin{equation}
\nabla\cdot\mathbf{u}=0,\label{eq:continuity}
\vspace{-1em}
\end{equation}

\begin{equation}
\frac{\partial\mathbf{u}}{\partial t}+\mathbf{u}\cdot\nabla \mathbf{u}=-\frac{1}{\rho}\nabla P+\nabla\cdot\left(\nu\nabla\mathbf{u}\right)-\nabla\cdot\mathbf{\tau},
\label{eq:RANS}
\end{equation}
where $\mathbf{u} = \left(u,v,w\right)$ is mean flow velocity, $t$ is time, $P$ is pressure, $\rho$ is density, $\nu$ is kinematic viscosity. $\mathbf{\tau}$ is the Reynolds stress tensor and is approximated by the RANS turbulence models. The standard $k-\epsilon$ turbulence model is used along with the wall function approach. Such a combination provides a balance between performance and computational efficiency~\cite{Li2021RANS}. Detailed mathematical expressions of the $k-\epsilon$ turbulence model can be found in previous studies~\cite{Launder1974}.

In this project, we consider five wind speed $[4,8,12,16,20]~m/s$ and for each speed we consider four wind directions $[0^\circ,90^\circ,180^\circ,270^\circ]$. Depending on the wind direction, the Dirichlet boundary condition of wind speed is applied to the corresponding upstream domain boundary surface, and the Neumann boundary conditions are applied to the downstream boundary surface. No slip boundary conditions are applied to all building surfaces and ground. The free-shear boundary conditions are applied to the top boundary domain and two-side boundary surfaces. A 5\% turbulence intensity is considered in the upstream boundary. The exact boundary conditions for turbulence quantities (i.e., $k, \epsilon$) are less of a concern in this study because the flow solutions are dominated by the turbulent wakes generated by the buildings. 

A Semi-Implicit Method for Pressure Linked Equations (SIMPLE) algorithm is used to solve the system of equations, i.e.,
Eqs. \ref{eq:continuity} and \ref{eq:RANS}. A second-order upwind scheme is used for the advection terms in the mean flow and turbulence equations. For the diffusion terms in the mean flow and turbulence equations, the second-order central-difference schemes are used. The simulations are considered as converged when the area-averaged turbulent kinetic energy (TKE) at the free surface becomes asymptotic (i.e., relative difference <0.1\%), and the scaled residuals of all variables are below $10^{-5}$.

\section{Experiments and Results}

In this section, we first introduce our training strategies, and explain our experiment set-up. Following this, we present the overall results and showcase example \mode{} flight paths.

\subsection{\mode{} Training Strategies}
We focus on optimizing two key aspects of \mode{} path planning: energy consumption and time cost.
In a dynamic setting such as a wind field, the two objectives can intrinsically conflict with each other. 
For instance, the optimization of time efficiency may come into conflict with energy conservation, particularly in the face of potential aggressive accelerations against wind resistance. To address this challenge, we train three distinct training strategies. The first strategy prioritizes minimizing energy consumption, the second concentrates on reducing travel time, and the third aims to strike a balance between both objectives.

\subsection{Experiment Set-up}

We report results about six wind fields, namely D0-4, D90-4, D180-4, D270-4, D0-8, and D0-12. 
The wind field's name consists of wind direction and speed, for example, D90-4 means that the angle between the wind direction and the positive X-axis is $90^\circ$ and the wind speed is $4~m/s$.

\nop{The experiment environment is a 3-D space containing buildings as obstacles, wind field vectors, origins and destinations. During the experiment, \mode{} starts in the origin and navigates to the destination. For our method, the RL agent tries to find the optimized flight paths under the three training strategies. Six wind field setups as D0-4, D90-4, D180-4, D270-4, D0-8, D0-12 are used for the three training strategies. The wind field named “D90-4” means that the angle between the wind direction and the positive X-direction is 90 degrees, and the wind boundary condition is 4 $m/s$. We setup three ODs for each training stages as Near-Distance, Mid-Distance, and Far-Distance. The OD and wind field setups apply to all the three training strategies.}

To validate our method, we compare our method with Dijkstra's algorithm~\cite{Dijkstra1959,Sniedovich2006}, which provides the optimal path over an urban area's network. 
Specifically, we create one graph for each wind field. These graphs' nodes represent the cells in the urban environment. We calculate the energy consumption and time cost for every possible movement between the cells (nodes) taken by \mode{} and add them into the graphs as the weights for the edges. Then we apply Dijkstra's algorithm to traverse the graphs, finding the theoretical optimal energy consumption and time cost for each graph and origin-destination pair.

\begin{figure*}
    \centering
    \includegraphics[width=\textwidth]{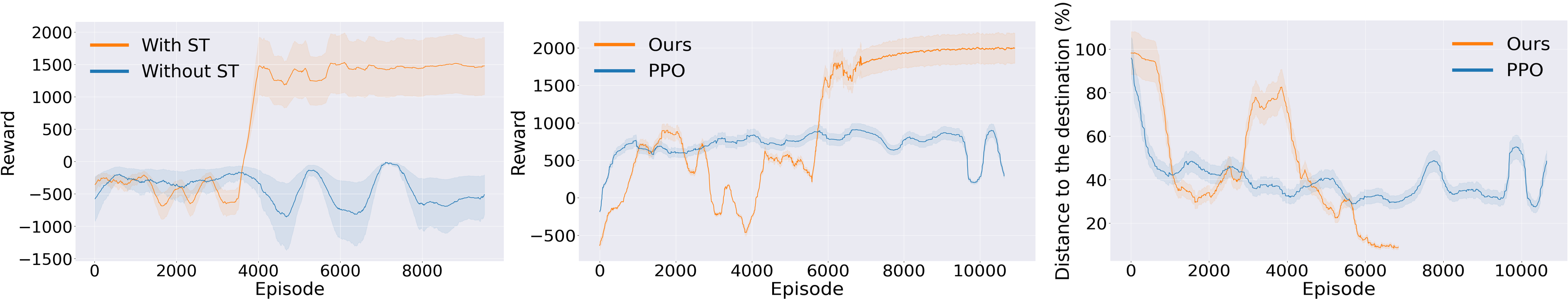}
    \caption{Training performance comparisons. LEFT: The addition of stage training (ST) enables significant improvement over its counterpart.
    MIDDLE: Our method starts to outperform vanilla PPO~\cite{Schulman2017} starting around the 6000$th$ episode. RIGHT: Our approach allows the \mode{} to reach its destination far earlier than vanilla PPO. These results demonstrate the effectiveness of our algorithm design. } 
    \label{fig:RL_Training_Effectiveness}
    \vspace{-.5em}
\end{figure*}

\subsection{Overall Results}
We first demonstrate the effectiveness of stage training as shown in Fig.~\ref{fig:RL_Training_Effectiveness} LEFT: an example learning curve shows a significant performance gain by incorporating the stage training. 
We then compare our approach and the vanilla PPO algorithm~\cite{Schulman2017}.
In Fig.~\ref{fig:RL_Training_Effectiveness} MIDDLE, we can see that our approach majorly outperforms PPO starting around the 6000$th$ episode. Lastly, Fig.~\ref{fig:RL_Training_Effectiveness} RIGHT shows our method can approach the destination around the 7000$th$ episode, while PPO still has around 30\% distance left to the destination around the 12000$th$ episode.

\begin{table*}[ht!]
\centering
\setlength{\tabcolsep}{5.75pt} 
\renewcommand{\arraystretch}{1.35} 
\begin{center}
\scalebox{1.06}{
    \begin{tabular}{|c|c|cccc|cccc|}
        \hline 
         Wind & OD & \multicolumn{4}{c|}{Energy ($kJ$)} & \multicolumn{4}{c|}{Time ($s$)} \\ 
        \cline{3-10} 
         Field & Index& Dijkstra-energy & Ours-energy & Ours-all & Ours' diff. (\%)& Dijkstra-time & Ours-time & Ours-all & Ours' diff. (\%)\\
        \cline{1-10}
        \multirow{3}{*}{D0-4} & 1 & 114.96& 122.78& 127.67& 3.83& 82.93& 87.47& 92.30 & 5.23\\
                              & 2& 107.33& 108.82& 114.21& 4.72& 80.53& 99.23& 100.18 & 0.95\\
                              & 3& 88.57& 88.9& 89.34& 0.49&64.56& 66.03& 70.34& 6.13\\
        \cline{1-10}
        \multirow{3}{*}{D90-4}  & 1& 113.51& 118.64& 122.25& 3.04&82.93& 87.79& 90.43& 2.92\\
                                & 2& 112.52& 113.37& 115.21& 1.60& 80.53& 87.37& 89.76&2.66\\
                                & 3& 91.12& 91.51& 97.89& 6.52&64.56& 65.34& 68.23&4.24\\
        \cline{1-10}
        \multirow{3}{*}{D180-4}  & 1& 107.12& 117.53& 124.52& 5.61&82.93& 87.42& 90.82&3.74\\
                                 & 2& 113.94& 116.95& 126.14& 7.29&80.53& 84.64& 87.65&3.43\\
                                 & 3& 85.19& 85.96& 86.73& 0.89&64.56& 65.35& 70.46&7.25\\
        \cline{1-10}
        \multirow{3}{*}{D270-4} & 1& 122.65& 125.21& 133.29&6.06& 82.93& 92.56& 97.22&4.79\\
         & 2& 103.8& 107.92& 110.72& 2.53&80.53& 97.52& 98.90&1.40\\
          & 3& 82.64& 85.62& 87.15&1.76& 64.56& 67.0& 68.12&1.64\\
        \cline{1-10}
        \multirow{3}{*}{D0-8}  & 1& 117.92& 129.36& 138.96& 6.91&82.93& 87.44& 90.31&3.18\\
        & 2& 108.66& 114.9& 115.54& 0.55&80.53& 92.16& 96.95&4.94\\
         & 3& 90.13& 94.59& 94.90&0.33& 64.56& 67.4& 69.56&3.11\\
        \cline{1-10}
        \multirow{3}{*}{D0-12}  & 1& 124.90& 140.45& 151.82&7.49& 82.93& 86.95& 92.43&5.93\\
         & 2& 109.55& 116.86& 121.32& 3.68&80.53& 85.02& 90.76&6.32\\
          & 3& 96.14& 104.42& 105.56& 1.08&64.56& 65.71& 68.89&4.62\\
        \hline
    \end{tabular}} 
\end{center}
\vspace{-8pt}
\caption{Energy and time costs of Dijkstra's algorithm and ours.
Each row represents a flight path between an OD pair within a wind field. Unpaired t-tests show no statistical significance found between Dijkstra's algorithm and our techniques. The differences between Ours-energy/time and Ours-all are also shown (Ours' diff.). All differences are under 10\% with most differences under 7\% for both energy and time costs. These results validate the effectiveness of our methods, particularly in achieving a balance between energy and time costs during \mode{} path planning. } 
\label{tab:results} 
\vspace{-1em} 
\end{table*}

\begin{table}[h]
\centering
\setlength{\tabcolsep}{5.75pt} 
\renewcommand{\arraystretch}{1.35} 
\begin{center}
\scalebox{.98}{
    \begin{tabular}{|c|cc|cc|cc|}
        \hline 
          & \multicolumn{2}{c|}{D0-4} & \multicolumn{2}{c|}{D90-4} & \multicolumn{2}{c|}{D180-4} \\ 
        \cline{2-7} 
          Method & Energy & Time  & Energy & Time & Energy & Time \\
        \cline{1-7}
         Ours-energy &  122.78&  109.01& 85.96&  77.88& 113.37& 144.6\\
         Ours-time   &  139.51&  87.47  &  108.2&  65.35 &  130.73& 87.37 \\
         Ours-all    & 127.67& 92.30 & 86.73& 70.46 & 115.21 & 89.76 \\
        \hline
    \end{tabular}} 
\end{center}
\vspace{-8pt}
\caption{Energy and time costs of our techniques in three wind fields. All results consistently show that Ours-all generates similar energy costs (with much less time costs) as of Ours-energy, and similar time costs (with much less energy costs) as of Ours-time. }
\label{tab:paths}
\vspace{-2.5em} 
\end{table}

Table~\ref{tab:results} shows the results of energy and time costs of various methods.
Each row represents a flight path within a wind field. 
The theoretical optimal values are generated by Dijkstra's algorithm focusing on energy-conservation (Dijkstra-energy) or time-saving (Dijkstra-time). 
The performances of our strategies focusing on energy-conservation (Ours-energy), time-saving (Ours-time), and balancing energy and time costs (Ours-all) are also shown. 

\begin{figure*}[ht]
    \centering
    \includegraphics[width=1\linewidth]{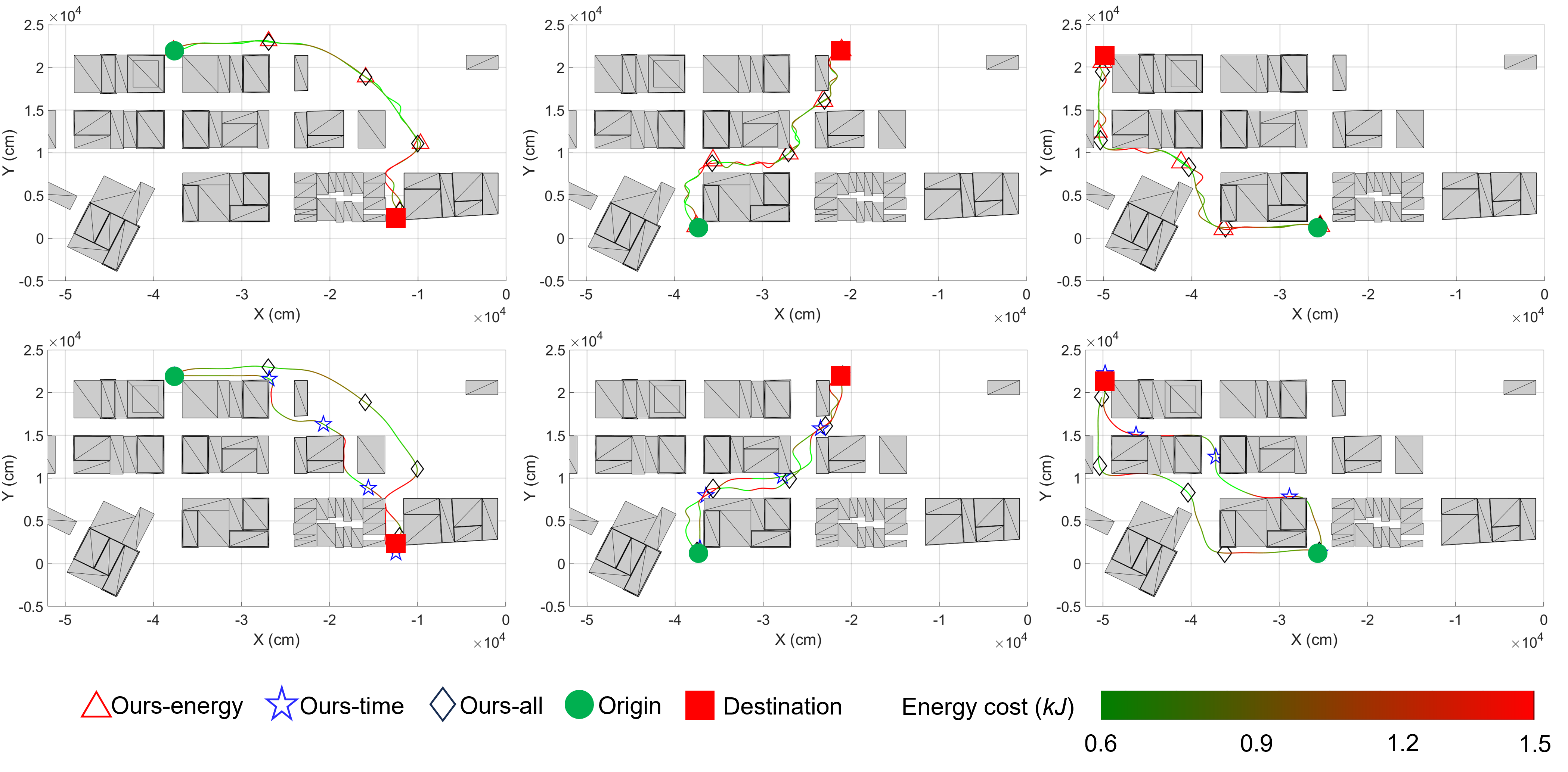}
    \vspace{-2em}
    \caption{Example \mode{} flight paths during wind field D0-4 (first column), D90-4 (second column), and D180-4 (third column). The subfigures in the first row show the comparisons of the paths from Ours-energy and Ours-all. The two sets of paths are highly similar indicating that Ours-all is effective in minimizing the energy consumption. The subfigures in the second row compare the paths from Ours-time and Ours-all. In these examples, Ours-all tends to opt for a different path, leveraging wind fields to achieve reduced energy consumption without significantly increasing travel times, as opposed to the paths chosen by Ours-time. }  
    \label{fig:12}
    \vspace{-2em}
\end{figure*}

To analyze the results, 
we perform a series of unpaired t-tests. 
We first compare the energy costs of Dijkstra-energy and Ours-energy, which results in the two-tailed $p=0.2894$, 
the difference in mean 5.17, and the 95\% confidence interval of this difference from -14.96 to 4.6.
By conventional criteria, this difference is considered to be not statistically significant, which indicates that our method generates similar results as of the theoretical optimal method in terms of energy cost.   
Next, we compare the energy costs of Ours-energy and Ours-all.
The two-tailed $p=0.4446$; the mean difference of Ours-energy and Ours-all is 4.41; the 95\% confidence interval of this difference is from -16.01 to 7.18. 
This difference is also not statistically significant, which indicates that Ours-all produces similar energy costs than Ours-energy, showing the effectiveness of Ours-all in energy conservation as well. 

About time cost, we first compare Dijkstra-time and Ours-time. 
The two-tailed $p=0.1028$, the difference in mean is 5.79, and the 95\% confidence interval of this difference is from -12.82 to 1.23. This difference is considered not statistically significant.
We then compare the time costs of Ours-time and Ours-all. The corresponding two-tailed $p=0.4049$, the difference in mean is 3.38, and the 95\% confidence interval of this difference is from -11.54 to 4.77. Again, this difference is considered not statistically significant. 
These results indicate that Ours-all is also effective in minimizing time cost, compared to Ours-time and Dijkstra-time.

\subsection{Example eVTOL Aircraft Flight Paths}

Fig.~\ref{fig:12} shows examples of \mode{} flight paths generated by our techniques. 
D0-4, D90-4, and D180-4 wind fields are adopted for the urban environment.
The black rectangles denote the paths by Ours-all, the cyan circles signify the paths by Ours-time, and the red triangles represent the paths by Ours-energy.
Each path is colored to show its energy cost.  
Ours-time yields more direct flight paths, aiming to reach the destination in the shortest possible time. In contrast, Ours-energy paths are more winding, exploiting wind fields in urban settings to save energy. 
Ours-all paths are trying to balance energy consumption and time cost at the same time.
The detail costs are listed in Table~\ref{tab:paths}. 
Overall, Ours-all consistently maintains comparable energy costs to Ours-energy while substantially decreasing travel time. Similarly, it exhibits comparable time costs compared to Ours-time, accompanied by a major reduction in energy consumption. 
These results suggest that our technique is not only effective in approximating the theoretically optimal values for energy and time efficiency but also proficient in balancing between these two objectives.

\section{Conclusion and Future Work} 

We introduce a novel method for \mode{} path planning in city-scale wind fields. 
Our method balances the energy conservation and time efficiency of \mode{} flight. We adopt reinforcement learning by significantly tailoring PPO~\cite{Schulman2017} to learn the optimal policy.  
To facilitate the learning process, we further adopt reward shaping and curriculum learning. 
To evaluate our method, we conduct comprehensive experiments comparing our method with Dijkstra’s algorithm -- the theoretically optimal approach for determining shortest paths in a weighted graph, where weights represent either energy or time cost. 
By varying wind fields and origin-destination pairs in our experiments, we show that our method produces similar results of the theoretically optimal approach.
Overall, our approach has proven to be highly effective and efficient in achieving a balance between energy and time efficiency during \mode{} path planning.
There are many future research directions. First, we plan to extend our technique to multi-\mode{} systems and study collaborative flight optimization in urban wind fields. 
Second, we will explore strategies in noise control of \mode{}. 
Third, various origin-destination demand patterns (as a result of different flight tasks) of \mode{} will be investigated.
Lastly, we would like to explore the joint study of air mobility and ground mobility in large-scale, mixed traffic settings~\cite{Li2017CityFlowRecon,Wang2023Intersection,Wang2024Privacy,Villarreal2024Eco}. 







\bibliographystyle{IEEEtran}
\bibliography{export}
\end{CJK}
\end{document}